\documentclass{article}
\usepackage{spconf,amsmath,graphicx,hyperref}

\usepackage{algorithm}
\usepackage{algorithmic}
\usepackage{amsmath,amssymb,amsthm}
\usepackage{cite}
\usepackage{color}
\usepackage{xcolor}
\usepackage{amsmath,amsthm,amssymb,amsfonts,bm}
\usepackage{textcomp}
\usepackage{multirow}
\usepackage{mathtools}

\usepackage{subcaption}

\usepackage{booktabs}   
\newenvironment{compactitem}{%
\begin{itemize}\setlength{\itemsep}{0pt}\setlength{\parskip}{0pt}\setlength{\parsep}{0pt}}{\end{itemize}}

\newtheorem{theorem}{Theorem}[section]

\theoremstyle{definition}
\newtheorem{definition}[theorem]{Definition}

\theoremstyle{remark}

\title{
Breaking Privacy in Federated Clustering: Perfect Input Reconstruction via Temporal Correlations
}

\newcommand{\threeauthblocks}[6]{%
  \name{%
    \begin{minipage}[t]{.32\linewidth}\centering #1\end{minipage}\hfill
    \begin{minipage}[t]{.32\linewidth}\centering #3\end{minipage}\hfill
    \begin{minipage}[t]{.32\linewidth}\centering #5\end{minipage}%
  }%
  \address{%
    \begin{minipage}[t]{.32\linewidth}\centering #2\end{minipage}\hfill
    \begin{minipage}[t]{.32\linewidth}\centering #4\end{minipage}\hfill
    \begin{minipage}[t]{.32\linewidth}\centering #6\end{minipage}%
  }%
}

\threeauthblocks
  {Guang Yang}%
  {Universit\'e Paris Cit\'e\\ France\\ g.1997.yang@gmail.com}%
  {Lixia Luo}%
  {Hunan University of Science and Technology\\ China\\ luolixia@hnust.edu.cn}%
  {Qiongxiu Li
  }%
  {Aalborg University\\ Denmark\\ qili@es.aau.dk}

\begin{document}
%
\maketitle

\begin{abstract}
Federated clustering allows multiple parties to discover patterns in distributed data without sharing raw samples. To reduce overhead, many protocols disclose intermediate centroids during training. While often treated as harmless for efficiency, whether such disclosure compromises privacy remains an open question. Prior analyses modeled the problem as a so-called  Hidden Subset Sum Problem (HSSP) and argued that centroid release may be safe, since classical HSSP attacks fail to recover inputs.  

We revisit this question and uncover a new leakage mechanism: temporal regularities in $k$-means iterations create exploitable structure that enables perfect input reconstruction. Building on this insight, we propose \emph{Trajectory-Aware Reconstruction (TAR)}, an attack that combines temporal assignment information with algebraic analysis to recover exact original inputs.   Our findings provide the first rigorous evidence, supported by a practical attack, that centroid disclosure in federated clustering significantly compromises privacy, exposing a fundamental tension between privacy and efficiency.

\end{abstract}


\section{Introduction}
\label{sec:intro}
Clustering is a core technique in unsupervised learning, widely used in data mining, anomaly detection, and recommendation systems~\cite{jain2010clustering}. In many practical scenarios, however, data are distributed across organizations and cannot be centralized due to privacy concerns. Federated clustering addresses this challenge by allowing multiple parties to jointly perform clustering while keeping raw data local~\cite{mcmahan2017communication}. Clients share intermediate statistics such as cluster sums or centroids, which are aggregated by a central server to update the model.    To prevent raw data disclosure, two main approaches have been explored. Differential privacy (DP) provides rigorous guarantees by injecting noise~\cite{stemmer2021locally,balcan2017differentially,su2017differentially}, but at the cost of accuracy. Cryptographic methods, including secure multiparty computation (SMPC)~\cite{Cramer2015,bunn2007secure,mohassel2019practical,kim2018privacy,erkin2009privacy,liu2014privacy,jiang2020efficient,li2022privacy,yuan2017practical} and homomorphic encryption (HE)~\cite{paillier1999public}, guarantee exact aggregation but incur high computational overhead. 

While these techniques protect raw data, the intermediate outputs of clustering algorithms—such as centroids, cluster assignments, and cluster sizes—also raise privacy concerns. As noted in~\cite{hegde2021sok}, it is difficult to rigorously determine whether releasing such information compromises privacy. Ensuring complete protection would require hiding every update, which increases complexity and reduces scalability. For efficiency, several protocols instead disclose aggregates such as centroids~\cite{erkin2009privacy,liu2014privacy,jiang2020efficient,li2022privacy} or assignments~\cite{sakuma2010large,yuan2017practical}. This raises a fundamental question: \emph{Does releasing intermediate statistics necessarily lead to privacy breaches?} 

A recent line of work sought to answer this question by framing federated $k$-means as an instance of the Hidden Subset Sum Problem (HSSP)~\cite{li2024privacy}. In this view, the disclosed cluster aggregates are interpreted as linear combinations of private data, and classical HSSP attacks such as Nguyen--Stern~\cite{nguyen1999hardness} and Coron--Gini~\cite{gini2022hardness} were shown to be ineffective due to the rank-deficiency of the underlying assignment matrix. This led to the prevailing belief that revealing intermediate centroids does not pose a significant privacy risk.  

In this paper, we revisit the problem and uncover a previously overlooked leakage mechanism. Specifically, we find that clustering iterations exhibit strong temporal regularities: many points remain in the same cluster across iterations, while others follow identical switching trajectories. These correlations induce algebraic structure that can be exploited for exact input reconstruction. Building on this insight, we introduce a new attack, \emph{Trajectory-Aware Reconstruction (TAR)}. TAR refines the assignment representation with temporal information and  applies an RREF-based test for recoverability, enabling both theoretical guarantees and concrete attacks that achieve perfect input reconstruction. We further validate TAR across diverse datasets to demonstrate its practical feasibility.  
Our key contributions are:
\vspace{-0.5\baselineskip}
\begin{compactitem}
    \item We provide, to the best of our knowledge, the first rigorous analysis showing that releasing intermediate centroids in federated $k$-means is fundamentally insecure.  
    \item We propose a novel attack TAR that exploits temporal correlations across iterations to reconstruct exact inputs.  
    \item We validate TAR extensively on diverse datasets, demonstrating high reconstruction success rates and highlighting the concrete risks of releasing intermediate centroids.  
\end{compactitem}

\section{Background and Related Work}
\label{sec:background}

\subsection{Hidden Subset Sum Problems}

The HSSP, defined over the ring of residues modulo positive integer $q$, provides a cryptographic framework for modeling 
privacy risks when aggregated sums are released. Formally, given a unknown binary matrix 
$\bm{A} \in \{0,1\}^{m \times n}$ and a hidden input vector $\bm{x} \in \mathbb{Z}^n$, 
the adversary observes the vector $\bm{h}$ satisfying
\[
\bm{h} \;=\; \bm{A} \bm{x} \pmod{q},
\]
and aims to recover both $\bm{A}$ and $\bm{x}$. 
Classical attacks on HSSP include the Nguyen--Stern algorithm~\cite{nguyen1999hardness} 
and its refinement by Coron and Gini~\cite{gini2022hardness}, which exploit lattice reduction techniques. 
These approaches crucially rely on the assumption that the binary entries of $\bm{A}$ 
are independent and identically distributed with probability $1/2$, so that 
$\mathbb{E}(\|\bm{a}_i\|_1)=m/2$ for each column $\bm{a}_i$. 
Under this random model---and in certain parameter regimes where $\bm{A}$ can be regarded as full rank or invertible---adversaries 
can, with high probability, reconstruct both $\bm{A}$ and $\bm{x}$.

\subsection{HSSP and Federated $k$-Means Clustering}
In federated clustering, clients share cluster statistics (e.g., sums/centroids) rather than raw data.  
Li and Luo~\cite{li2024privacy} observed that releasing intermediate cluster sums in federated $k$-means can be formalized as a $d$-dimensional HSSP instance over $\mathbb{Z}$, where $d$ denotes the feature dimension of the input samples.

Let $\bm{X} \in \mathbb{Z}^{n \times d}$ denote the concatenation of local data values, 
where $n$ is the total number of samples. 
Let $\bm{C} \in \mathbb{Z}^{kT \times d}$ collect the released cluster sums over $T$ iterations, 
with $k$ clusters per iteration, and let 
$\bm{W} \in \{0,1\}^{kT \times n}$ be the corresponding assignment matrix 
indicating which points belong to which clusters at each iteration.
These quantities satisfy the linear system
\[
\bm{C} = \bm{W}\bm{X}.
\]
Because such $\bm{W}$ instances are often rank-deficient and highly structured, prior work concluded that standard HSSP attacks (Nguyen--Stern; Coron--Gini) are ineffective on these instances.

While casting federated $k$-means as an HSSP instance is insightful, we find that it overlooks key properties of the algorithm. The assignment matrix $\bm{W}$ is not random but exhibits fixed sparsity, repeated columns, and dependencies created by iterative updates. Although its rank deficiency has been interpreted as a sign of privacy, such underdetermined systems can still permit recovery once structural information is taken into account. Furthermore, successive iterations are temporally correlated: many points remain fixed in clusters while others follow repeated switching patterns, introducing algebraic dependencies across time. Taken together, these factors suggest that privacy in federated $k$-means should be assessed under its structured and temporally correlated assignment distribution, rather than through random HSSP models.

\section{Proposed approach}
\label{sec:tar}
We now proceed to introduce the proposed approach that exploits the temporal  correlations for input reconstruction.
\subsection{System Representation}
In practice, many samples never change clusters, while others follow identical switching trajectories. As a result, the raw assignment matrix $\bm{W}$ contains numerous duplicate columns. We first deduplicate repeated columns to obtain
\begin{equation}\label{eq:dedup-system}
\bm{C} = \bm{W}^\star \bm{Y},
\end{equation}
where $\bm{W}^\star\in\{0,1\}^{kT\times n^\star}$ contains unique trajectories, 
and $\bm{Y} = [\bm{y}_i^\top]_{i=1}^{n^\star}$ aggregates variables sharing the same path.

\subsection{RREF-based Leakage Criterion}
We further analyze leakage through the reduced row echelon form (RREF). Form the augmented matrix
\[
[\, \bm{W}^\star \mid \bm{C} \,] \;\sim\; [\, \bm{W}^\star_{\mathrm{rref}} \mid \bm{C}_{\mathrm{rref}} \,],
\]
where the right-hand side is obtained by Gaussian elimination in RREF.
\begin{definition}[Successful Attack]  \label{def.1}
An attack is \emph{successful} if some row of $\bm{W}^\star_{\mathrm{rref}}$ contains exactly one nonzero entry, in which case the corresponding variable $\bm{y}_j$ is uniquely determined.  Note that this criterion guarantees \emph{perfect reconstruction} as the recovered $\bm{y}_j$ matches the original input exactly (e.g., zero $L_2$ error).
\end{definition}

Unlike classical HSSP analyses that target full recovery of $\bm{X}$, here even the recovery of a single $\bm{y}_j$ constitutes a privacy breach. For instance, in the one-dimensional example
\[
\begingroup
\setlength{\arraycolsep}{3pt}\renewcommand{\arraystretch}{0.95}
\left[\begin{array}{@{}cccc@{}}\!1&1&1&0\\1&1&0&1\\0&0&1&1\!\end{array}\right]\bm{Y}
=\left[\begin{array}{@{}c@{}}c_1\\c_2\\c_3\end{array}\right]
\;\Rightarrow\;
\left[\begin{array}{@{}cccc|c@{}}\!1&1&0&0&\!\tfrac{c_1+c_2-c_3}{2}\\0&0&1&0&\!\tfrac{c_1-c_2+c_3}{2}\\0&0&0&1&\!\tfrac{c_2-c_1+c_3}{2}\end{array}\right]_{.}
\endgroup
\]
Though the system admits infinitely many solutions, the third and fourth variables are uniquely determined, demonstrating leakage despite underdeterminacy.

\subsection{Proposed Trajectory-Aware Reconstruction Attack}
We now describe how to construct $\bm{W}^\star$ from the information disclosed in $k$-means clustering. Since cluster assignments change simultaneously across all dimensions, the construction can be reduced to a single column. For notation simplicity, we assume $d=1$ in \eqref{eq:dedup-system}, but the results generalize to arbitrary dimension (see Section \ref{sec:experiments} for experimental results across various dimensions $d$). We now have 
\[
\bm{c} \;=\; \big[\, (\bm{c}^{(1)})^\top, \; (\bm{c}^{(2)})^\top, \; \dots, \; (\bm{c}^{(T)})^\top \,\big]^\top 
\;\in\; \mathbb{R}^{kT}.
\]
where $\bm{c}^{(t)} \in \mathbb{Z}^k$ is the vector of cluster sums at iteration $t$.

\textbf{Step 1: Differences and Index Sets.}  
The first step is to capture how cluster sums evolve across iterations.  
For each iteration $t$, 
define
\begin{equation*}
\begin{aligned}
\bm{\Delta}^{(t)} &\coloneqq \bm{c}^{(t+1)}-\bm{c}^{(t)}, \\
\mathcal{I}^{(t)}_+ \coloneqq \{j : \bm{\Delta}^{(t)}_j > 0\},& 
\;\;\text{ and }\;\; 
\mathcal{I}^{(t)}_- \coloneqq \{j : \bm{\Delta}^{(t)}_j < 0\}.
\end{aligned}
\end{equation*}
These sets indicate which clusters gain or lose points between iterations.  
We then assign a column budget
\[
r_t = \max\{|\mathcal{I}^{(t)}_+|,\;|\mathcal{I}^{(t)}_-|\},
\]
which determines the number of templates required at time $t$.

\textbf{Step 2: Template Blocks.}  After obtaining the index sets $\mathcal{I}^{(t)}_+$, $\mathcal{I}^{(t)}_-$ and the column budget $r_t$,  we construct the template blocks $\bm{B}^{(t)}, \bm{D}^{(t)} \in \{0,1\}^{k\times r_t}$ of $\mathcal{I}^{(t)}_-$ and $\mathcal{I}^{(t)}_+$, respectively, by applying the \emph{fill-one rule} given as follows.

The fill-one rule of an index set $\mathcal{I}\subseteq \{1,\dots,k\}$ and a target column budget $r$  to construct a template block $\bm{M}\in\{0,1\}^{k\times r}$ contains: 1)  For each $j\in\mathcal{I}$, assign a $1$ in a distinct column of $\bm{M}$, ensuring every index in $\mathcal{I}$ appears at least once; 2)  If $r > |\mathcal{I}|$, fill the remaining columns by randomly repeating rows from $\mathcal{I}$ (random completion);
3)  If $\mathcal{I}=\emptyset$, set $\bm{M}=\mathbf{0}$.

\textbf{Step 3: Recursive Combination.}  
We then combine the per-iteration templates to form a global matrix capturing all switching patterns.  
At $t=1$, initialize
\begin{align} \label{W_1}
\bm{W}^{(1)}=
\begin{bmatrix}
\bm{B}^{(1)}\\[2pt] \bm{D}^{(1)}
\end{bmatrix}\in\{0,1\}^{2k\times r_1}.
\end{align}
For $t \ge 2$, let $\bm{W}^{(t-1)}_{\mathrm{end}}$ denote the last $k$ rows of $\bm{W}^{(t-1)}$. The update is
\begin{align} \label{W_t}
\bm{W}^{(t)}=
\begin{bmatrix}
\bm{W}^{(t-1)} & \operatorname{Rep}_{t-1}( \bm{B}^{(t)})\\[4pt]
\bm{W}^{(t-1)}_{\mathrm{end}} & \bm{D}^{(t)}
\end{bmatrix}.
\end{align}
After $T$ iterations we obtain $\bm{W}'=\bm{W}^{(T-1)}\in\{0,1\}^{kT\times n_W}$, which represents all switching trajectories observed across the run.  

\textbf{Step 4: Stationary Points.}  
The recursive construction only accounts for samples that switch clusters at least once.  
In practice, many points remain in the same cluster throughout.  
To represent these stationary samples, we extend $\bm{W}'$ with identity blocks:
\vspace{-5pt}
\begin{equation}\label{eq:W-extension}
\bm{E}
= \Big[\ \overbrace{\bm{I}_k \mid  \bm{I}_k \mid  \cdots \mid  \bm{I}_k}^{T\ \text{times}}\ \Big]^{\!\top}
\in \{0,1\}^{kT\times k},
\end{equation}
and set
\[
\widetilde{\bm{W}} = \big[\, \bm{W}' \mid \bm{E} \,\big].
\]

\textbf{Step 5: Final Deduplication.}  
Finally, note that some of the trajectories represented in $\widetilde{\bm{W}}$ may coincide, producing duplicate columns.  
We therefore merge identical columns to obtain the final iteration-record matrix $\bm{W}^\star$, which compactly represents all possible trajectories and forms the basis of our reconstruction attack.

\begin{algorithm}[t]
\footnotesize
\caption{Proposed Trajectory-Aware Reconstruction}
\label{alg:iter-record-final}
\begin{algorithmic}[1]
  \STATE \textbf{Input:} cluster sums $\{\bm{c}^{(t)}\}_{t=1}^T$, number of clusters $k$
  \STATE \textbf{Output:} $\bm{W}^\star\in\{0,1\}^{kT\times n^\star}$
  \STATE $\bm{c}\leftarrow \operatorname{vec}([\bm{c}^{(1)},\dots,\bm{c}^{(T)}])$ \hfill (stacked observables)
  \FOR{$t=1$ \TO $T-1$}
    \STATE $\bm{\Delta}^{(t)}\leftarrow \bm{c}^{(t+1)}-\bm{c}^{(t)}$
    \STATE $\mathcal{I}^{(t)}_+\leftarrow\{j:\bm{\Delta}^{(t)}_j>0\}$;\quad $\mathcal{I}^{(t)}_-\leftarrow\{j:\bm{\Delta}^{(t)}_j<0\}$
    \STATE $r_t\leftarrow \max\{|\mathcal{I}^{(t)}_+|,|\mathcal{I}^{(t)}_-|\}$ \hfill (per-iteration column budget)
  \ENDFOR
  \STATE Construct $\bm{B}^{(1)}, \bm{D}^{(1)}$ using the fill-one rule
  \STATE Set $\bm{W}^{(1)}$ via Eq.~\eqref{W_1}
  \FOR{$t=2$ \TO $T-1$}
    \STATE Construct $\bm{B}^{(t)}, \bm{D}^{(t)}$ using the fill-one rule
    \STATE Let $\bm{W}^{(t-1)}_{\mathrm{end}}$ be the last $k$ rows of $\bm{W}^{(t-1)}$
    \STATE Update $\bm{W}^{(t)}$ via Eq.~\eqref{W_t}
  \ENDFOR
  \STATE $\bm{W}' \leftarrow \bm{W}^{(T-1)}$ \hfill (samples that switched at least once)
  \STATE Construct $\bm{E}$ via Eq.~\eqref{eq:W-extension}
  \STATE $\widetilde{\bm{W}} \leftarrow [\, \bm{W}' \mid \bm{E} \,]$
  \STATE Merge identical columns of $\widetilde{\bm{W}}$ to obtain $\bm{W}^\star$
  \STATE \textbf{Return} $\bm{W}^\star$
\end{algorithmic}
\end{algorithm}

\subsection{Algorithmic Summary}
Algorithm~\ref{alg:iter-record-final} summarizes the proposed input reconstruction attack. After obtaining $\bm{W}^\star$, we then use the RREF criterion to execute exact input reconstruction and calculate the attack success rate.

\begin{table}[!t]
\centering
\small
\setlength{\tabcolsep}{4pt} 
\caption{Attack success rates of the proposed TAR for synthetic datasets for three dimensions $d=1,2,3$ under various iteration numbers.}
\label{tab:iter-stats-wide}
\resizebox{\columnwidth}{!}{%
\begin{tabular}{c | c c c | c c c | c c c}
\toprule
\multirow{2}{*}{\textbf{Iter.}} 
& \multicolumn{3}{c|}{\textbf{1D}} 
& \multicolumn{3}{c|}{\textbf{2D}} 
& \multicolumn{3}{c}{\textbf{3D}} \\
& \textbf{All} & \textbf{Passed} & \textbf{Passed rate}
& \textbf{All} & \textbf{Passed} & \textbf{Passed rate}
& \textbf{All} & \textbf{Passed} & \textbf{Passed rate} \\
\midrule
2  & 211 & 0   & 0.0\%  & 168 & 0   & 0.0\%  & 147 & 0   & 0.0\%  \\
3  & 265 & 228 & 86.0\% & 266 & 226 & 85.0\% & 285 & 239 & 83.9\% \\
4  & 189 & 135 & 71.4\% & 217 & 143 & 65.9\% & 225 & 139 & 61.8\% \\
5  & 104 & 46  & 44.2\% & 156 & 86  & 55.1\% & 159 & 74  & 46.5\% \\
6  & 43  & 7   & 16.3\% & 75  & 13  & 17.3\% & 82  & 19  & 23.2\% \\
7  & 36  & 9   & 25.0\% & 44  & 7   & 15.9\% & 49  & 7   & 14.3\% \\
8  & 19  & 5   & 26.3\% & 21  & 4   & 19.0\% & 26  & 4   & 15.4\% \\
9  & 10  & 0   & 0.0\%  & 22  & 1   & 4.5\%  & 14  & 2   & 14.3\% \\
10 & 123 & 2   & 1.6\%  & 31  & 1   & 3.2\%  & 13  & 0   & 0.0\%  \\
\midrule
\textbf{Total} 
& \textbf{1000} & \textbf{432} & \textbf{43.2\%}
& \textbf{1000} & \textbf{481} & \textbf{48.1\%}
& \textbf{1000} & \textbf{484} & \textbf{48.4\%} \\
\bottomrule
\end{tabular}}
\vspace{-5pt}
\end{table}

\section{Experimental validations and discussion}
\label{sec:experiments}

We evaluate the proposed TAR attack on synthetic datasets and two real-world benchmarks: 
the Iris~\cite{fisher1936use} dataset and the Olivetti Faces~\cite{samaria1994olivetti} dataset. 
Together, these experiments span controlled, interpretable, and high-dimensional scenarios. In each case, we measure the percentage of runs in which TAR achieves perfect input reconstruction, 
i.e., the RREF criterion certifies uniqueness and the reconstructed inputs match the originals exactly.
We consider two disclosure regimes: (i) the full trajectory, where centroids from all iterations are available; and (ii) a truncated-trajectory setting, where only the last few $L$ centroid updates are disclosed. 
The latter reflects the fact that in $k$-means, the cluster assignments tend to stabilize at later iterations, while only a few input points continue switching. This simplifies the structure of $\bm{W}^\star$ and may still suffice for recovery.
Overall success rates are summarized in Table~\ref{tab:overall-success}, with dataset-specific patterns discussed below.

\subsection{Synthetic Data}
The first is synthetic datasets for assessing leakage under controlled conditions. 
For $n=20$ integer samples uniformly drawn from $[0,50]$ with dimensions $d=1,2,3$, 
we run $k$-means clustering with $k=4$ clusters and random initializations for $T=10$ iterations.
Each experiment is repeated 1000 times. 
Table~\ref{tab:iter-stats-wide} reports success rates of 43.2\% ($d=1$), 48.1\% ($d=2$), and 48.4\% ($d=3$) when all iterations' centroid information is used. 
Restricting disclosure to the last $L=6$ iterations actually \emph{increases} the rates to 45.2\%, 51.4\%, and 48.7\%, respectively, 
showing that late-stage updates alone preserve, and can even amplify, temporal leakage.

\subsection{Iris and Olivetti Faces Datasets}
\begin{figure}[t]
    \centering

  \begin{subfigure}{0.15\linewidth}\includegraphics[width=\linewidth]{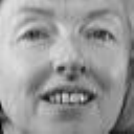}\end{subfigure}\hspace{0.01\linewidth}
  \begin{subfigure}{0.15\linewidth}\includegraphics[width=\linewidth]{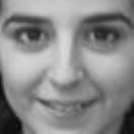}\end{subfigure}\hspace{0.01\linewidth}
  \begin{subfigure}{0.15\linewidth}\includegraphics[width=\linewidth]{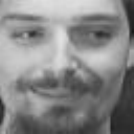}\end{subfigure}\hspace{0.01\linewidth}
  \begin{subfigure}{0.15\linewidth}\includegraphics[width=\linewidth]{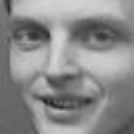}\end{subfigure}\hspace{0.01\linewidth}
  \begin{subfigure}{0.15\linewidth}\includegraphics[width=\linewidth]{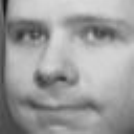}\end{subfigure}\hspace{0.01\linewidth}
  \begin{subfigure}{0.15\linewidth}\includegraphics[width=\linewidth]{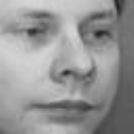}\end{subfigure}

  \vspace{0.6ex}

  \begin{subfigure}{0.15\linewidth}\includegraphics[width=\linewidth]{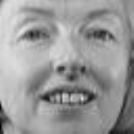}\end{subfigure}\hspace{0.01\linewidth}
  \begin{subfigure}{0.15\linewidth}\includegraphics[width=\linewidth]{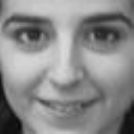}\end{subfigure}\hspace{0.01\linewidth}
  \begin{subfigure}{0.15\linewidth}\includegraphics[width=\linewidth]{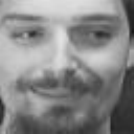}\end{subfigure}\hspace{0.01\linewidth}
  \begin{subfigure}{0.15\linewidth}\includegraphics[width=\linewidth]{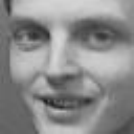}\end{subfigure}\hspace{0.01\linewidth}
  \begin{subfigure}{0.15\linewidth}\includegraphics[width=\linewidth]{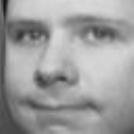}\end{subfigure}\hspace{0.01\linewidth}
  \begin{subfigure}{0.15\linewidth}\includegraphics[width=\linewidth]{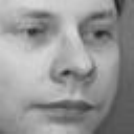}\end{subfigure}

    \caption{Olivetti Faces: originals (top row) vs. reconstructions (bottom row) obtained from $k$-means clustering. The L2 norm between reconstructed and original input images is 0, namely perfect (exact) input reconstruction.}
    \label{fig:olivetti_reconstruction}
    \vspace{-5pt}
\end{figure}

We evaluate TAR on both the Iris dataset ($n=150$, $d=4$, $k=3$ clusters) and the high-dimensional Olivetti Faces dataset (400 grayscale images, $64 \times 64$, $d=4096$).  
For the Iris dataset, we perform 1000 runs with up to $T=10$ iterations. TAR achieves perfect reconstruction in 44.6\% of runs when using full trajectory data, with success rates rising to 51.0\% when only the last $L=6$ iterations are disclosed.  
This mirrors the synthetic results, confirming that even truncated information remains sufficient for exact recovery.  
In the case of the Olivetti Faces dataset, TAR achieves a 77\% success rate across 500 runs, with reconstructions visually indistinguishable from the originals ($L_2=0$), as shown in Fig.~\ref{fig:olivetti_reconstruction}.  
These results demonstrate that temporal leakage is not confined to small, low-dimensional datasets but extends to more complex, high-dimensional images, reinforcing the broad applicability of the proposed TAR approach.

\subsection{Discussion}
\noindent\textbf{Answer to the central question.}  
As posed in Sec.~\ref{sec:intro}: \emph{does releasing intermediate centroids necessarily lead to privacy breaches?}  
Our results provide the first rigorous empirical evidence that the answer is \emph{yes}.  
TAR achieves certified \emph{perfect input reconstruction} with substantial success rates across synthetic, tabular, and image data.

These findings establish that intermediate centroids, often disclosed to improve efficiency and reduce communication, expose severe and quantifiable privacy risks.  
They characterize the \emph{baseline (lower-bound) leakage} under unprotected centroid disclosure.  
In practice, mitigating this risk requires additional defenses (e.g., DP, secure multiparty computation, or homomorphic encryption), but such techniques inevitably reintroduce the accuracy–efficiency trade-off.

\begin{table}[t]
\centering
\footnotesize
\caption{Overall success rates of the proposed TAR for 5 datasets with both full and truncated ($L=6$) settings. }
\label{tab:overall-success}
\begin{tabular}{lccc}
\toprule
\textbf{Dataset} & \textbf{Runs} & \textbf{Full} & \textbf{Truncated} \\
\midrule
Synthetic (d=1)   & 1000 & 43.2\% & 45.2\% \\
Synthetic (d=2)   & 1000 & 48.1\% & 51.4\% \\
Synthetic (d=3)   & 1000 & 48.4\% & 48.7\% \\
Iris (d=4)  & 1000 & 44.6\% & 51.0\% \\
Olivetti Faces (d=4096)  & 500  & 77.0\% & 77.0\% \\
\bottomrule
\end{tabular}
\end{table}

\section{Conclusion}
\label{sec:conclusion}

We introduced the TAR, a new attack that leverages temporal assignment regularities in federated $k$-means. 
By combining a recursive iteration-record matrix with an RREF-based leakage criterion, we formally and empirically demonstrated that intermediate centroid disclosure enables exact recovery of private inputs. 
Our results provide the first rigorous evidence that federated clustering, as commonly deployed, is fundamentally insecure. 
Future research should focus on developing privacy-preserving clustering methods that reconcile this risk with the efficiency requirements of real-world federated systems.

\newpage
\bibliographystyle{IEEEtran}

\bibliography{refs}

\end{document}